# Breaching the Bottleneck: Evolutionary Transition from Reward-Driven Learning to Reward-Agnostic Domain-Adapted Learning in Neuromodulated Neural Nets

Solvi Arnold[1], Reiji Suzuki[2], Takaya Arita[2] and Kimitoshi Yamazaki[1]

[1] Department of Mechanical Systems Engineering, Shinshu University
[2] Graduate School of Informatics, Nagoya University
s_arnold@shinshu-u.ac.jp

## Abstract

Advanced biological intelligence learns efficiently from an information-rich stream of stimulus information, even when feedback on behaviour quality is sparse or absent. Such learning exploits implicit assumptions about task domains. We refer to such learning as Domain-Adapted Learning (DAL). In contrast, AI learning algorithms rely on explicit externally provided measures of behaviour quality to acquire fit behaviour. This imposes an information bottleneck that precludes learning from diverse non-reward stimulus information, limiting learning efficiency. We consider the question of how biological evolution circumvents this bottleneck to produce DAL. We propose that species first evolve the ability to learn from reward signals, providing inefficient (bottlenecked) but broad adaptivity. From there, integration of non-reward information into the learning process can proceed via gradual accumulation of biases induced by such information on specific task domains. This scenario provides a biologically plausible pathway towards bottleneck-free, domain-adapted learning. Focusing on the second phase of this scenario, we set up a population of NNs with reward-driven learning modelled as Reinforcement Learning (A2C), and allow evolution to improve learning efficiency by integrating non-reward information into the learning process using a neuromodulatory update mechanism. On a navigation task in continuous 2D space, evolved DAL agents show a 300-fold increase in learning speed compared to pure RL agents. Evolution is found to eliminate reliance on reward information altogether, allowing DAL agents to learn from non-reward information exclusively, using local neuromodulation-based connection weight updates only. Code available at github.com/aislab/dal.

## Introduction

Learning is an indispensable aspect of intelligence. Perhaps the most extensively studied form of biological learning is learning from reward, as a central focus of study in behaviourist psychology. Thorndike's (1898) *Law of Effect*, followed by Skinner's (1938) *Operant Conditioning* (OC) formalised how reward modifies behaviour. However, as it became evident that the reward characteristics of stimuli alone are insufficient to explain biological learning, the behaviourist paradigm fell out of favour. Advanced biological intelligence learns readily from various types of non-reward information. Reward plays a special role, but we are not constrained to learning from reward alone. This versatility allows us to learn efficiently and robustly even when reward signals are scarce.

Within AI, the introduction of learning algorithms may have produced more dramatic progress than any other innovation. The learning abilities of AI systems, in particular Neural Networks (NNs) are inspired by learning in biological species. However, mainstream NN learning algorithms resemble biological learning only superficially. Training by means of error backpropagation requires definition of an error (loss) signal quantifying the output quality, which is then propagated backward through the NN to obtain gradients for updating synaptic weights. Neither the backpropagation procedure nor the existence of an explicit error signal is biologically justified. These discrepancies do not just strain the biological metaphor, but also impose limitations on AI learning that biological learning evades. We will focus on Reinforcement Learning (RL), as it maps onto biological learning more readily than other mainstream learning algorithms, enabling behaviour learning without explicit instruction. RL is, in essence, OC recast and refined as an algorithm, and as such it inherits OC's shortcomings. The central quality measure driving the RL process is (cumulative discounted) reward. All association of stimuli (observations) to responses (actions) is controlled by this quantity. In RL the dependence on reward coincides with the requirement of an explicit scalar quality measure for backpropagation: loss signals for connection weight updates are calculated from the scalar reward signal. In essence, RL turns behaviour learning into a game of hot and cold. We will refer to this reliance on a scalar quality measure as the *reward bottleneck*.

These discrepancies between biological and artificial learning play a key role in the *efficiency gap* between them. The *results* of AI learning can be spectacular, but the learning process itself is slow and arduous, requiring far larger amounts of data and energy than biological learning processes. This makes it infeasible for AI systems to learn effectively from limited real-world experience, limiting their flexibility and adaptivity.

These observations raise the following questions. How does biological intelligence evade the reward bottleneck? How does it manage to exploit non-reward information for learning? Can we turn the answers to these questions into more efficient AI learning algorithms?

Our computational study will rely heavily on *neuromodulation* (NM). Biological brains actively modify their own synapses on basis of local neural activity and internally generated modulatory signals. Because these signals can derive from any information flowing into the network, NM-based learning is naturally free of information bottlenecks. Along with its biological plausibility, this makes it particularly suitable for modelling learning from diverse types of information.

### Contributions

1) A demarcation of Domain-Adapted Learning (DAL) abilities as a category of learning abilities in biological species that is poorly accounted for by AI training methods.
2) A theory explaining the evolvability of DAL via catalysis by reward-driven learning (RDL).
3) A computational model based on this theory wherein DAL is evolved using RL and neuromodulation.
4) A direct comparison between a mainstream RL algorithm and an evolved DAL agent, revealing a substantial advantage in learning efficiency in the DAL agent.

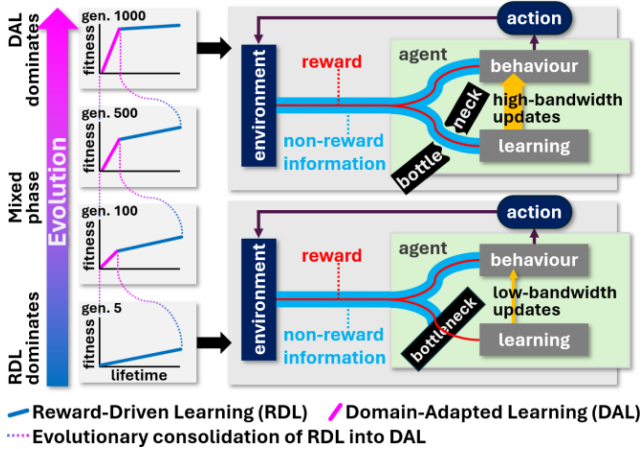

Figure 1. DAL evolution. RDL provides consistent learning processes over generations, which evolution consolidates into efficient DAL by integrating non-reward information into the learning process, thereby eliminating the reward bottleneck.

## Theory

We start by observing that learning from reward is, in a sense, easy. Action sequences that yield high/positive rewards should be repeated, actions that yield low/negative rewards should be avoided. We can state this much without any knowledge about the task to be learned. However, if we are presented with a signal $X$ unaccompanied by a reward value, what can we learn? In this case, to learn from $X$, we will need an understanding of what we are aiming to learn, and how $X$ pertains to that goal. That is, we need to make *assumptions* about the *task domain*.

The concept of task domains is central for understanding the conditions that allow learning from non-reward information. We loosely define a task domain as a set of task instances with the same underlying structure, but different surface features. Natural languages, for example, share fundamental structures, but differ in their surface features. Hence language acquisition is an example of a task domain, composed of specific languages as its *task instances*. Learners of different languages acquire different verbal behaviour, but their learning processes are similar. Similarly, we can view the task of learning to locomote as a task domain. Individuals of a given species differ in their physical characteristics, so each individual is presented a different instance of the locomotion learning task. Different individuals acquire subtly different gaits, but the learning process and acquired functionality are similar. In the language acquisition case, an individual trying to learn to communicate will need to discover regularities in co-occurrences between specific vocalisations and objects or events. Discovering and exploiting such regularities can be done just fine in absence of reward. Humans readily acquire language regardless of whether explicit reinforcement is provided. Similar arguments can be made for other task domains. If we "know" what we are trying to learn (i.e. make assumptions about the task domain), then we can exploit non-reward information to learn it. Below we refer to learning abilities that exploit assumptions about the task domain as Domain-Adapted Learning (DAL).

Biological evolution and the field of AI differ in their willingness to make such assumptions. Assumptions constrain learning abilities to specific task domains. In other words, making assumptions implies stepping away from algorithmic generality. The field of AI greatly values generality in learning algorithms. For AI, much of the value of learning lies in that it saves us from having to hand-design behaviours for many different tasks, so the prospect of designing learning abilities for many different task domains is unappealing. Moreover, the existence of general-purpose learning algorithms shadows the utility of domain-specific ones. Biological evolution is less concerned with algorithmic elegance or generality. Biological species evolve in specific niches that require acquisition of specific skills and behaviours, with a premium on learning speed. If it is advantageous to have multiple domain-specific learning abilities adapted to multiple survival-essential task domains, we can expect evolution to prefer plurality over parsimony.

Here we see a trade-off between generality and information bandwidth. Domain-specificity affords assumptions, which afford learning from non-reward information, which affords more efficient learning. However, the example set by biology also shows that generality and domain-adapted learning are not mutually exclusive, *if we are willing to stack two levels of adaptation*. Evolution is a general-purpose adaptation process producing domain-specific learning abilities.

How does DAL evolve? Unfortunately, evolutionary search for plastic neural architectures that instantiate particular DAL abilities is a *needle in a haystack* problem. At least two factors complicate such evolution. One is the large search space presented by evolutionary search for learning abilities in general. Evolution of neural architectures already presents a large search space, and the space of possible update dynamics over that space is larger still. A second complicating factor is specific to DAL. Reward-driven learning (RDL) is obtained when a correlation is established between received reward quantities and changes in the probability of the preceding actions. In contrast, DAL involves more subtle and complex contingencies between a larger bandwidth of incoming information and the behaviour modifications that specific information therein should elicit. Consequently, DAL evolution requires a fine synergy between behaviour and learning. Agents have to behave in just such a way as to elicit useful information, while also, through evolutionary coincidence, having the right neural mechanism in place to respond to exposure to that information with an advantageous behaviour update. Dependence on such unlikely confluences makes DAL challenging to evolve artificially.

These complications raise the question of what allowed DAL to evolve readily in biological evolution. Here we propose that RDL is a powerful catalyst for evolution of DAL. Consider a population of agents learning by RDL, as approximated by OC/RL. The reward bottleneck limits learning efficiency, but RDL does consistently provide agents with progress towards fitter behaviour. Then a mutation allows some non-reward information to leak through the bottleneck and perturb the learning process. This will induce a bias that can either be beneficial or detrimental. If the induced bias is beneficial, the leak will be retained by evolutionary selection. Accumulation of such leaks gradually integrates more and more non-reward information into the learning process, opening up the reward bottleneck and reducing the species' reliance on explicit reward signals from the environment. In this scenario, RDL is responsible for discovering useful behaviours to learn (*what* to learn), and for providing consistent learning dynamics over generations. This allows evolution of DAL to proceed as gradual efficiency improvement over those learning dynamics (*how* to learn). Gradual consolidation of RDL into DAL frees up time and energy for additional RDL, allowing for discovery of further behaviour improvements. Repeated cycles of discovery and consolidation produce abilities for rapid acquisition of increasingly complex behaviour on the task domains important to the species. Figure 1 illustrates this evolutionary transition.

We aim to demonstrate this scenario in an evolving population of NNs, using RL as an approximation of RDL, and neuromodulation as basis for DAL evolution. Our evolutionary scenario can be considered a variant of the Baldwin effect (Baldwin, 1896, Simpson, 1953), with the genetic assimilation of acquired traits replaced with genetic assimilation of mechanisms for acquiring certain classes of acquired traits efficiently.

## Related Work

### Domain-Adapted Learning in Nature

Above we introduced the concept of DAL as a catch-all for forms of learning that exploit non-reward information in a domain-specific manner. Such forms of learning are not usually grouped into a single category, but studies of specific examples abound. Certain examples from the end of the behaviourist era are particularly instructive, as they are specifically designed to escape explanation in terms of RDL. Garcia et al. (1955) found that rats readily associate consumption of water with a specific flavour to subsequent sickness. In typical conditioning scenarios, reward must follow action within a brief time window for an association between the two to take hold, but rats make this particular association even if there is a substantial delay between the consumption of the water and the onset of sickness (Garcia et al., 1966). Furthermore, even if simultaneous presentation of gustatory, auditory, and visual stimuli precedes the onset of sickness, rats still selectively associate the gustatory stimulus with subsequent sickness (Garcia & Koelling, 1966). This shows that not just the value of the reward, but also the characteristics of the stimuli (including that of the reward itself) play a major role in determining which associations are formed. In other words, non-reward information biases the learning process. We see that this bias latches on to a basic fact of rat physiology: ingestion of substances is a likely culprit for subsequent sickness, more so than bells or flashing lights are.

A more advanced example is rats' ability to learn to navigate mazes in absence of reward, as demonstrated in e.g. detour mazes (Tolman & Honzik, 1930). Here, rewards are not obtained until *after* learning is completed, so such learning cannot be explained as a consequence of reinforcement. Furthermore, rats and other species are known to possess neural circuitry dedicated to the acquisition of spatial representations (Moser et al., 2008). Such circuitry can be understood as an adaptation to the task domain of spatial navigation learning.

We already mentioned language acquisition as an example of a task domain. In language acquisition, we observe highly efficient learning in a task setting with information-rich observations, but very sparse rewards. This strongly suggests that the reward bottleneck is bypassed. Indeed, behaviourist attempts to explain language acquisition in terms of OC (Skinner, 1957) were unconvincing (Chomsky, 1959).

These examples illustrate that learning abilities fitting the definition of DAL are diverse, common, and broadly recognised in the literature.

### Neuromodulated Learning in Artificial Agents

As mentioned, our computational study will rely on neuromodulation (NM) as its basis for learning. NM-based learning extends Hebbian update logic with modulatory signals from elsewhere in the network, modelling the effects of endogenously controlled local neuromodulator concentrations. Various attempts at evolving NM-based learning in NNs are found in the literature on Evolved Plastic Artificial Neural Networks (Soltoggio et al., 2018). An early example of NN learning on basis of endogenously generated signals is found in Nolfi & Parisi (1993). Soltoggio et al. (2007, 2008) modelled such signals explicitly after biological neuromodulation, and demonstrated their effectiveness in reward-based scenarios. The suitability of NM-based method for evolving DAL specifically has flown mostly under the radar, but exceptions are found in Arnold et al. (2012, 2013, 2014) where forms of learning that qualify as DAL are evolved using modulatory mechanisms. In the task scenarios of Nolfi & Parisi (1993) (foraging) and Yamato et al. (2022) (meta-memory), reward exists but is not visible to the NNs, meaning that learning necessarily employs different types of information. Hebbian plasticity without modulatory signals has also been explored in reward-free scenarios (Najarro & Risi, 2022, Ferigo et al., 2022). Note that even when learning is evolved in reward-based scenarios, the fact that evolution occurs in a specific scenario makes it possible for non-reward information to contribute to the learning process.

As discussed above, evolving learning from scratch is challenging, and despite its theoretical potential, research in this area still remains mostly limited to interesting proofs of concept. Recognising the absence of a smooth fitness gradient as a major hurdle, some have explored *novelty search* (Lehman & Stanley, 2011) to improve evolvability of NM-based learning in reward-based scenarios (Risi et al., 2010). Optimisation of plasticity with a global scalar modulatory signal using back-propagation instead of evolution has been explored for such scenarios as well (Miconi et al., 2018). However, the biologically plausible scenario of RDL catalysing evolution of more efficient NM-based learning remains unexplored.

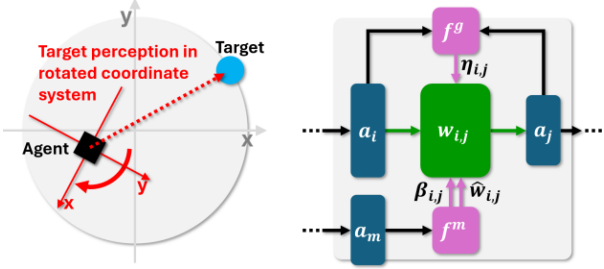

Figure 2. Left: Task design. Agents see positions of randomly placed targets in a randomly rotated local coordinate system. Right: schematic of our neuromodulation mechanism.

## Model

### Task Setup

We design a task domain as follows. Each *trial* starts with the agent placed at the origin of the 2D plane, and a target placed at a distance of 1 from the origin in a random direction. The task for the agent is to approach the target as closely as possible in $T_{trial} = 10$ timesteps. At every timestep, the agent observes the position of the target relative to its own current position as a 2D vector, and outputs a 2D motion vector as its action. Figure 2 (left) illustrates the task. Action vectors are rescaled into the unit circle and divided by $T_{trial}$, resulting in a maximum movement speed of $1/T_{trial}$ per timestep. At time $t = T_{trial}$, the agent receives a reward $r = 1 - d$, where $d$ is the remaining distance between the agent and the target. Consequently, the theoretical maximum score for a trial is 1. Note that sitting still at the start position yields a reward of zero, while running off in arbitrary directions yields negative reward on average.

To necessitate individual learning, we randomise the orientation of the coordinate system in which the observation of the target is given, while the orientation of the world coordinate system in which actions are performed remains fixed. Consequently, the agent cannot employ a fixed mapping from observations to actions, and has to acquire a suitable mapping by means of learning instead. The collection of tasks produced by this randomisation constitutes our task domain, and individual randomisations are task instances.

Note that there are two levels of randomisation at work. Orientation of the observation coordinate system is randomised per task instance, while the target position is randomised at the start of each trial within each task instance.[1] Performing well on this task requires untangling these randomisation levels. The combination of performed actions and resulting changes in the perceived target position implicitly provides information that makes such untangling possible.

### Reinforcement Learning Algorithm

Numerous RL algorithms exist, but only a subset is suitable for our purpose. We aim to evolve agents with a fully neural architecture, that learns while it acts within a continuous task space. This results in the following requirements. (1) Continuous state and action domains. (2) Online learning. (3) Direct action output from the NN substrate. (4) No complex support structures external to the NN (e.g. replay buffers). A2C, a popular and performant variant of A3C (Mnih et al., 2016), meets these requirements. Note however that it is not the current SOTA on benchmark RL tasks. Comparisons to algorithms incompatible with the above requirements are outside the scope of this work. For compatibility with evolving NN architecture and integration with NM weight updating logic, we reimplement A2C from scratch. We adopt the default hyperparameter settings from the *Stable-Baselines3* (SB3) reference implementation (Raffin et al., 2021), except for the update interval (number of steps over which updates are aggregated), which we match to the trial length.[2] To assess the difficulty level of our task, we train from scratch using RL only, with an NN architecture identical to the initial architecture used in our evolutionary experiments (detailed below). The learning process (averaged over 10 task instances) can be found in Figure 4g. It takes A2C over 3000 trials to converge at high performance, indicating that the task is sufficiently challenging to provide a meaningful testbed.

### Neural Network Architecture

We refer to *sets of neurons* as *columns*. An *activatory projection* is a set of synapses connecting one column to another. Each activatory projection connects all neurons in its *pre-synaptic* column to all neurons in its *post-synaptic* column. A *modulatory projection* takes input from a column but terminates on an activatory projection, and outputs weight updates for the set of synapses in its target projection. The choice to let modulation operate at the level of individual synapses instead of neurons is motivated by the observation that cholinergic modulation (involved in e.g. associative learning) can selectively act on subsets of a neuron's synapses (Fellous et al., 2016).

A typical A2C agent contains separate critic and actor NNs. We adopt a functionally equivalent single-net architecture that splits into two branches after the input column. Each branch consists of one hidden column and one output column. One branch outputs a distribution over the action space and the other provides a scalar state value estimate, replicating the output of the usual A2C setup. Internal architecture may change over the course of evolution, but input and output columns are always retained. All hidden columns consist of $N_{hidden} = 8$ neurons, using the hyperbolic tangent activation function.

The NN receives as input the current state observation (2D vector indicating relative position of the target in the current perception coordinate system), the preceding state observation, and the agent's preceding action (2D vector in world coordinate system). Note that the reward signal is not included in the input (it acts on the agent via the RL update logic exclusively). Following A2C, the action distribution output by the NN consists of means and standard deviations for each element of the action vector. Actions are chosen by random sampling from this distribution. Broad distributions result in random exploratory behaviour, while narrow distributions result in focused near-deterministic action choice. Output activation functions are identity for action means and state value estimates, and the exponential function for action standard deviations.

---

[1] Approaching arbitrary targets can be implemented as a fixed mapping, so target randomisation alone does not change the task instance.
[2] Our task uses a fixed trial length of $T_{trial} = 10$, and (non-zero) reward is only received at the end of each trial. An update interval of 10 makes more sense than the SB3 default of 5 in this case, and was found to produce marginally better performance.

Activatory projections are defined in the genotype by their initial weight matrix and a projection-local RL learning rate for weight updates. This local learning rate makes it possible for evolution to suppress RL selectively in specific projections. Local RL learning rates are set to 1 in the initial population. Local update rates are multiplied by a global RL learning rate, which is initialised to 0 in the initial population.

Modulatory projections have internal structure (Fig. 2, right), designed to provide high flexibility for mapping activation patterns in the modulating columns to weight updates. Let $Q_m$ be a modulatory projection modulating an activatory projection $P_{i,j}$ between pre-synaptic column $i$ and post-synaptic column $j$, with column $m$ as the modulating column. $Q_m$ consist of two MLPs, $f^m$ and $f^g$, with one hidden column each, of size $2N_{hidden}$ and $N_{hidden}$, respectively.[3] $f^m$ maps the activation pattern $a_m$ of the modulating column $m$ to a matrix $\widehat{w}_{i,j}$ of target weights and a matrix $\beta_{i,j}$ of update rates, each equal in size to weight matrix $w_{i,j}$ of projection $P_{i,j}$.

$$(\widehat{w}_{i,j}, \beta_{i,j}) = f^m(a_m) \qquad (1)$$

The activation function of the output layer of $f^m$ is linear for $\widehat{w}_{i,j}$, and for $\beta_{i,j}$ it adds 0.5 and clamps to the [0,1] interval. $f^g$ maps activation patterns $a_i$ and $a_j$ of the pre- and post-synaptic columns to a scalar update rate $\eta_{i,j}$.

$$\eta_{i,j} = f^g(a_i, a_j) \qquad (2)$$

The activation function of the output layer of $f^g$ adds 0.5 and clips to the [0,1] interval. Eq. 3 gives the NM update rule for weight matrix $w_{i,j}$.

$$\Delta w_{i,j} = \eta_{i,j} \cdot \beta_{i,j} \odot (\widehat{w}_{i,j} - w_{i,j}) \qquad (3)$$

The synaptic weights in $w_{i,j}$ step towards the target set by $\widehat{w}_{i,j}$, with the size of the step determined by the projection-level update rate $\eta_{i,j}$ and the synapse-level update rates given by $\beta_{i,j}$. The intended function of $\beta_{i,j}$ is to act as a mask for selecting subsets of synapses to modify. A synapse's weight can be left as-is by pushing the corresponding element of $\beta_{i,j}$ to 0. The primary function of $f^g$ is to assess the progress of the learning process at $P_{i,j}$ from the activation patterns on its pre- and post-synaptic columns, and suppress further updates when finished. (Continued updates with similar target weights would keep weights fixed at those targets, which would block RL from further modifying the weights.) Rates $\eta_{i,j}$ and $\beta_{i,j}$ only affect modulation-based updates. A2C handles its own dynamic synapse-local update rates via the RMSProp optimiser.

Modulatory projections are defined in the genotype by the set of connection weight matrices specifying $f^m$ and $f^g$, plus a priority gene determining the update order when multiple projections terminate on the same activatory projection.

Recall that the reward signal is not part of the NN's input. Furthermore, the modulating column $m$ for a modulating projection is selected from the input and hidden columns, thereby excluding the value estimate output. NM projections thus do not have access to the reward signal or the estimate thereof. However, the local information available to modulatory projections can include, directly or indirectly, any information present in the NN's input. Hence the information flow into the NM-based update system is free of bottlenecks.

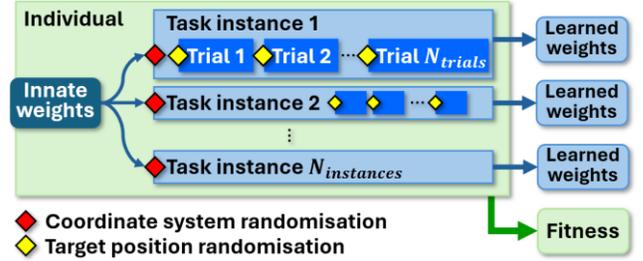

Figure 3. Agent evaluation process.

### Evolution

We evolve a population of 100 agents for 1500 generations, using a parent pool of size 25 with an elite pool of size 5. Each agent is evaluated on $N_{instances} = 64$ task instances, starting from its innate state on each instance. We use the same batch of task instances for all individuals within a generation, and generate a new batch of task instances for every generation. Each agent learns on each task instance for $N_{trials} = 50$ trials. Fig. 3 illustrates the evaluation process for one agent.

Mutations can add or remove projections and (indirectly) columns.[4] Projection weight matrices are mutated by adding uniformly sampled noise of random magnitude to random subsets of their connection weights. Global and local RL update rates are perturbed using random uniform noise.

We introduce two *guided mutation* operators. These use the result of the parent individual's learning processes to perform non-random mutations. Guided mutation on an activatory projection shifts the projection's weight matrix towards the mean of the set of weight matrices obtained by the parent on its set of task instances, by a random amount. Half of all activatory projection weight matrix mutations are guided.

Guided mutation on a modulatory projection attempts to optimise the MLP weight matrices comprising it such that it will reproduce the local effects of learning in the parent network. Consider again a modulatory projection $Q_m$ modulating activatory projection $P_{i,j}$, with column $m$ as the modulating column. For brevity, let $T = N_{trials} \cdot T_{trial}$ (i.e. the number of timesteps spent on each task instance) and $H = N_{instances}$. Let $a_k^{t,h}$ denote the activation vector of column $k$ at timestep $t \in [1 \ldots T]$ in task instance $h \in [1 \ldots H]$, and let $w_{i,j}^{t,h}$ denote the weight matrix of projection $P_{i,j}$ at timestep $t$ in task instance $h$. If time series $a_{m,i}^{1\ldots T, 1\ldots H}$ contains sufficient information for predicting the local effects of learning at $P_{i,j}$, then we can find MLP weight matrices for $Q_m$ that, when exposed to $a_{m,i}^{1\ldots T, 1\ldots H}$, will cause $P_{i,j}$ to approach a set of weight matrices $\tilde{w}_{i,j}^{1\ldots H}$ that are functionally approximately equivalent to $w_{i,j}^{T, 1\ldots H}$, by optimising the weights of $Q_m$ so as to minimise loss $l$.

---

[3] The hidden columns of the MLPs comprising modulatory projections could as well be considered columns of the encompassing network. Encapsulating them as part of the modulatory projection is motivated by the mutation operators we define over modulatory projections.
[4] Each NN implicitly contains 7 columns, but only those contained in a circuit connecting the input column to an output column are part of its computational graph. Consequently, projection insertions/deletions can effectively cause columns to appear/disappear.

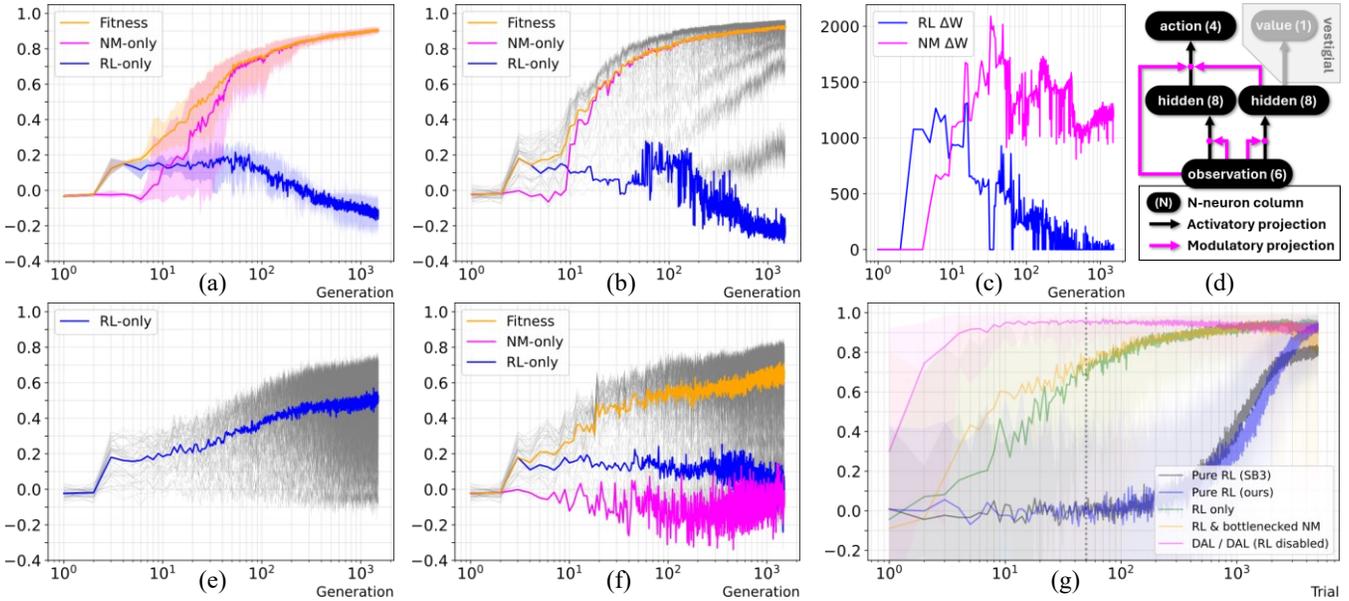

Figure 4. (a) Evolutionary dynamics over 5 runs of the main model. Fitness scores are of the focal individual. Lines are averages over all runs. Shaded areas indicate the range of observed values. (b) Evolutionary dynamics of selected run. Fitness scores are of the focal individual. Grey lines show reward obtained in the first, second, …, last trial of the lifetime. (c) Evolutionary dynamics of the sum total connection weight modification by RL and NM in the selected run of the main experiment. (d) Architecture of the final focal individual from the selected run. (e) RL-only run. (f) A run with an RL-like information bottleneck imposed on NM. (g) Comparison of learning speeds of various agents. Averages over 32 task instances. Shaded regions indicate the range of observed values over all task instances. Note that all x-axes are in logscale. Runs in (b), (e) and (f) use the same seed, leading to similar initial dynamics.

$$l = \sum_{h=1}^{H} \left[ s(r^h) \cdot \sum_{t=1}^{T} \left\| z^j(\tilde{w}_{i,j}^{t,h} \cdot a_i^{t,h}) - z^j(w_{i,j}^{T,h} \cdot a_i^{t,h}) \right\| \right] + \tau \cdot \eta_{i,j}^{t,h}$$

$$s(r^h) = \frac{r^h - min(r^{1\ldots H})}{max(r^{1\ldots H}) - min(r^{1\ldots H})} \quad (4)$$

Here $\tilde{w}_{i,j}^{t,h}$ is the weight matrix at projection $P_{i,j}$ after updating by $Q_m$ on basis of activation patterns $a_{m,i}^{1\ldots t,h}$, $z^j$ is the activation function of column $j$, $r^h$ is the mean reward collected by the parent over the last five trials on task instance $h$, and $\eta_{i,j}^{t,h}$ is the synapse level update rate at timestep $t$ in task instance $h$. System parameter $\tau = 10^{-5}$ imposes a small cost on updates. Weighing task instances by collected reward prioritises replication of learning processes on task instances where learning was more effective. Recall that $Q_m$ operates on non-reward information, and note that the loss definition rewards learning speed (the sooner $\tilde{w}_{i,j}^{t,h}$ approximates the behaviour of the parent's *final* weight matrix $w_{i,j}^{T,h}$, the smaller the loss). Optimising the weight matrices comprising $Q_m$ under loss $l$ thus explicitly implements the concept of efficiency improvement via integration of non-reward information into the learning process. Note that $w_{i,j}^{T,h}$ can be the result of RL alone or any mixture of RL and NM-based learning in the parent, allowing for repeated cycles of discovery and improvement (cf. Fig. 1, left).

We optimise $Q_m$ using signSGD (Bernstein et al., 2018) with a dynamically adjusting update rate. Activation patterns from 48 of the parent's 64 task instances are used for optimisation and the remaining 16 for determining when to terminate the process. Optimisation success is contingent on whether the modulating column provides suitable and sufficient information. When it does not, optimisation stalls early and is cut short. We apply this optimisation whenever a mutation inserts a new modulatory projection, and in half of all modulatory projection weight matrix mutations.

## Results

### Evolutionary Dynamics

We ran the evolutionary process 5 times. Figure 4a shows aggregate results, and Figure 4b an example run selected for further analysis. We plot the following quantities. *Fitness*: mean reward over all trials comprising the lifetime of an unmutated clone of the fittest individual from the preceding generation (referred to as the *focal individual* below). This provides an unbiased performance indicator. *NM-only*: fitness obtained when the focal individual is evaluated with RL disabled. As the reward signal is invisible to NM-based learning, positive NM-only fitness is indicative of learning from non-reward information (i.e. DAL). Difference between regular fitness and NM-only fitness quantifies RL's contribution to the learning process. *RL-only*: fitness obtained when the focal individual is evaluated with NM-based weight updates disabled. The difference between regular and RL-only fitness quantifies NM's contribution to the learning process. Plots are in log-scale, because most transitions of interest occur during early generations. Next we look at the development of the roles of RL and NM in our selected run (Fig. 4b). Figure 4c quantifies the amount of weight modification performed by RL and NM in the focal individual at each generation in this run. We distinguish 4 phases:

1) [Generation 3] Appearance of RDL. Note that all it takes to "evolve" RDL is to mutate the global RL learning rate to a suitable positive value. Along with a rise in regular fitness and RL-only fitness (Fig. 4b), we see a rapid rise in weight modification by RL (Fig. 4c).
2) [Generation 5] Appearance of NM in a supportive role. NM-based weight modification rises, and a gap appears between regular fitness and RL-only fitness, indicating that NM has started to contribute to the learning process. NM-only fitness remains low, indicating that NM in isolation does not produce meaningful learning yet.
3) [Generation 10~] Evolution of NM-based learning (DAL). NM-only fitness starts to rise, indicating the appearance of purely NM-based learning ability. From there, both regular fitness and NM-only fitness rise rapidly, with a steadily decreasing gap (indicating RL's contribution to the learning process) in between. This phase matches the evolutionary cycle of RDL progress being consolidated into more efficient DAL, freeing up RDL to discover new progress, which then gets consolidated into DAL, freeing up RDL to discover new progress, etc. NM-based weight modification rises while RL-based modification falls.
4) [Generation 100~] Convergence & Obsolescence of RL. As the gap between regular and NM-only fitness narrows, evolutionary progress levels off, and converges at high fitness. The differential between regular and NM-only fitness evaporates, indicating that learning is now entirely driven by NM working on non-reward information. RL-based weight modification drops to zero by the end of the run, with occasional small non-zero blips.

As is evident from Figure 4a, all runs produced similar dynamics. We observe some variation in the timing of evolutionary transitions, but DAL evolves rapidly and consistently. In Figure 4c, weight modification due to RL shows a rise-and-fall pattern characteristic of Baldwinian plasticity dynamics. Fig. 4d shows the NN architecture of the focal individual of the final generation. We observe that the original activatory connectivity has been retained in this run, but has become subject to heavy modulation. Because RL-based weight modification has been eliminated in this agent, the value output neuron is vestigial.

For comparison, we run RL-only and NM-only versions of the evolutionary experiment.[5] Figure 4e shows the RL-only run. Global and local RL learning rates and innate connection weights evolve as in the main experiment. This setup resembles MAML (Finn et al., 2017) and the approach of Fernando et al. (2018), in that adaptation to the task set happens primarily by optimisation of initial weights. While this is a form of DAL (learning is adapted to the task domain), learning ability remains constrained by the RL framework. We see that evolution improves learning speed, reaching a lifetime fitness of about 0.52, but falling short of NM-based DAL. Evolution in the NM-only experiment flatlined around zero (graph omitted for being uninformative). We can infer from the main experiment that RL-free high-fitness solutions exist within the search space, but in absence of RDL, evolution failed to discover any within 1500 generations. This is consistent with the idea that discovering DAL from scratch is a needle in a haystack problem that can be circumvented with guidance from RDL.

To see how a 1D information bottleneck constrains learning, we ran an experiment where RL and NM are both present, but NM projections are limited to using the value output neuron as the modulating neuron. Evolution might modify the content of the information passed through this neuron (hence this is not strictly a *reward* bottleneck), but the bandwidth of the signal remains fixed to 1D. Figure 4f shows the resulting evolutionary trajectory. NM works its way into the learning process (as evidenced by the difference between regular fitness and RL-only fitness), but remains constrained to a supporting role. We see some fitness improvement compared to the RL-only run, but otherwise the runs are similar. This suggests that circumvention of the information bottleneck is essential to the high efficiency obtained by NM in the main experiment.

### Comparison of Learning Efficiency

We compare learning ability between final-generation focal individuals from the main experiment, the RL-only run, the bottlenecked-NM run, and a pure (non-evolved) A2C agent. Figure 4g plots performance (mean collected reward over 32 task instances) over 5000 trials. The trial axis is plotted in log-scale. As previously established, pure RL takes over 3000 trials to converge to high performance on our task. The evolved RL-only agent improves much faster initially, reaching rewards of about 0.7 in 50 trials (agent lifetime during evolution, dotted line). The agent with RL and bottlenecked NM learns slightly faster still, reaching about 0.75 in 50 trials. The DAL agent reaches mean rewards over 0.9 in 5 trials, and converges in 10. Its performance over trials 10 to 50 exceeds final performance of pure RL by a small margin. Note that it obtains a mean reward of 0.3 at the end of the first trial. This is not due to fit innate behaviour, but due to substantial learning occurring within the first trial. The RL learning rate is zero in this agent, so disabling RL has no effect on its performance. Recall that the reward signal is invisible to the NM update logic, due to its exclusion from the state observation. The DAL agent thus outperforms the pure RL agent in terms of learning speed by a factor of over 300, while learning exclusively from non-reward information, using local NM-based update logic only.

---

[5] For clarity: in the main set of experiments, evaluations of the focal individual with RL or NM disabled are performed for purpose of analysis only and do not affect evolution. The comparative experiments are set up and evolved with only RL or only NM present in the system.

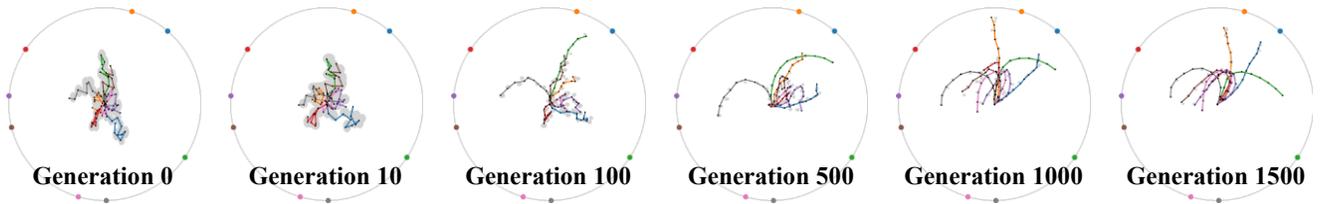

Figure 5. Development of the learning strategy over the course of evolution. We retrieve the focal individual for each plotted generation and run it on a fixed set of 8 task instances, for *one* trial each. Colours distinguish task instances. Coloured dots on the perimeter indicate target positions. Coloured lines indicate agent trajectories. Grey ellipses indicate action distributions (mean ± 1SD).

### Analysis of Learning Process

Next we analyse how the learning process changes over the course of evolution. Figure 5 shows the first trial of the learning process of the focal agent from a selection of generations, on a test set of 8 task instances. In early generations, first-trial trajectories extend into arbitrary directions, and action distributions are relatively broad. This suggests a random trial-and-error learning style, with significant use of the RL algorithm's random action sampling mechanism. In late generations, first-trial trajectories initially all extend in similar directions, before curving towards the target position. Action distributions have narrowed, indicating reduced reliance on random action sampling. This suggests a transition away from trial-and-error, towards focused experimentation for gathering the specific information the learning system has evolved to expect and exploit.

## Discussion & Conclusions

We demonstrated that 1) reward-driven learning can catalyse evolution of DAL, 2) NM can instantiate DAL by integrating non-reward information into the learning process, 3) this can improve learning efficiency and 4) avoid dependence on reward signals. These results support the viability of our theory of DAL evolution, and the practical feasibility of evolving DAL in NNs. We saw evolution transfer a learning process from one substrate (RL) to another (NM), while compressing it into a fraction of its original time cost. NM is inherently free of information bottlenecks, making its potential advantages over backpropagation particularly salient for scenarios where learning from reward or error alone is suboptimal.

Comparison between domain-agnostic RL and DAL with an evolutionary history on the task domain may seem unfair. It may seem unreasonable to expect RL to keep up from such a disadvantaged starting point. It is equally unreasonable to expect general-purpose AI learning algorithms to keep up with biological learners in the world those biological learners evolved in. The comparison illustrates the *cost of generality* that such algorithms incur, and the need to consider this cost if we hope to realise efficient learning in artificial systems.

It may seem that foregoing generality for efficiency would lead us away from human-like general intelligence. But despite the overlap in terminology, the mathematical generality of AI learning and the apparent generality of human intelligence have little in common. We learn on novel domains not by reverting to blank slates, but by applying cognitive tools from familiar domains to novel ones. This is perhaps most evident when we reason by analogy. To understand the versatility of human intelligence, we should understand how this cognitive toolbox comes about, and how we transfer it across domains. Studying the evolution of DAL is a step toward the former.

The ability to learn without relying on reward or error signals has relevance to the topic of *autonomy*. A learner that is hardcoded to maximise (minimise) a reward (error) signal supplied by the environment is controlled by that environment. Learning mechanisms that are less reliant on reward and endogenously control how information modifies behaviour may be better suited to produce some semblance of self-determination.

The evolutionary chronology of RDL and DAL in our theoretical scenario is consistent with the fact that RDL is observed in comparatively primitive species. It is reasonable to assume that such learning appeared relatively early in evolution, and was available for catalysing evolution of DAL. However, our model also includes some biologically implausible elements. We used backpropagation-based RL to model RDL. A more complete model would start by evolving NM-based RDL with a sufficient level of generality. The use of "off-the-shelf" RL did, however, not prevent us from obtaining performant backpropagation-free NM-based learning abilities. Another biologically implausible element is the use of guided mutations. In future work we plan to explore scenarios where the optimisation of projection weights is left fully to selection over random mutations. A third divergence from biology is that we withheld reward signals from NM-based learning, to facilitate analysis of the contribution of non-reward information.

How well our computational model extends to task domains of higher dimensionality and complexity remains to be seen. We have obtained positive results on a 3D version of the navigation task, and are exploring more complex domains.

How can this work contribute to smarter AI learning? If all we need is fixed behaviour on a specific task instance, the cost of evolving DAL exceeds that of training a general-purpose learner from scratch. However, if we hope to make learning an active part of a deployed system's *functionality*, pre-evolving DAL for its deployment domain could provide the ability to learn efficiently from limited experience in the real world, perhaps even in absence of explicit reward structures. Every real-world application presents a battery of potential assumptions and information streams that could, in theory, be integrated into such systems' learning abilities.

## Acknowledgments

This work was supported by JSPS KAKENHI Grant Number JP23K11262. We thank Yusuke Yamato and Mahiro Kato for insightful discussions during the development of this work.